\documentclass[lettersize,journal,10pt]{IEEEtran}
\usepackage{amsmath,amsfonts}
\usepackage{algorithmic}
\usepackage{algorithm}
\usepackage{array}
\usepackage[]{footmisc}
\usepackage{textcomp}
\usepackage{stfloats}
\usepackage{url}
\usepackage{verbatim}
\usepackage{graphicx}
\usepackage{cite}
\usepackage{multirow}
\usepackage{tabularx}
\usepackage{booktabs}
\usepackage{lscape}
\usepackage{longtable}
\usepackage{multicol}
\usepackage{subcaption}
\usepackage{float}
\newcommand{\ourmethod}{{$\textsc{REFINE-AF}$}}
\usepackage{times}
\usepackage{latexsym}
\usepackage[colorlinks=true, citecolor=blue]{hyperref}
\usepackage{url}
\usepackage{float}
\usepackage{makecell}
\usepackage{adjustbox}
\usepackage{amsmath}
\usepackage{xcolor}

\begin{document}

\title{REFINE-AF: A Task-Agnostic Framework to Align Language Models via Self-Generated Instructions using Reinforcement Learning from Automated Feedback}

\author{
    \large\IEEEauthorblockN{Aniruddha Roy, 
                      Pretam Ray, 
                      Abhilash Nandy,
                      Somak Aditya,
                      Pawan Goyal} \\
        \normalsize\IEEEauthorblockA{Indian Institute of Technology, Kharagpur. \\
        Email: \{aniruddha,pretam,abhilash\}@iitkgp.ac.in, \{saditya,pawan\}@cse.iitkgp.ac.in}
}


\maketitle

\begin{abstract}
Instruction-based Large Language Models (LLMs) have proven effective in numerous few-shot or zero-shot Natural Language Processing (NLP) tasks. However, creating human-annotated instruction data is time-consuming, expensive, and often limited in quantity and task diversity. Previous research endeavors have attempted to address this challenge by proposing frameworks capable of generating instructions in a semi-automated and task-agnostic manner directly from the model itself. Many of these efforts have relied on large API-only parameter-based models such as GPT-3.5 (175B), which are expensive, and subject to limits on a number of queries.
This paper explores the performance of three open-source small LLMs such as LLaMA 2-7B, LLama 2-13B, and Mistral 7B, using a semi-automated framework, thereby reducing human intervention, effort, and cost required to generate an instruction dataset for fine-tuning LLMs. Furthermore, we demonstrate that incorporating a Reinforcement Learning (RL) based training algorithm into this LLMs-based framework leads to further enhancements. Our evaluation of the dataset reveals that these RL-based frameworks achieve a 
substantial improvements in 63-66\% of the tasks compared to previous approaches. 
\end{abstract}


\section{Introduction}
Significant progress has been achieved in recent NLP research towards empowering large language models to comprehend instructions provided in natural language 
However, the process of gathering and annotating such instruction data is labor-intensive and costly, and often limited in quantity, diversity, and creativity. 

Several automated or semi-automated methods 
\cite{zhou2023large,ye2023guess,singh2023explaining,honovich2022instruction,wang2023selfinstruct,li2024selfalignment} have been developed for instruction generation.  \cite{zhou2023large} introduces Automatic Prompt Engineer (APE), drawing inspiration from program synthesis and human prompt engineering for automatic instruction generation and selection. \cite{ye2023guess} proposes flipped learning for language models, training the model to generate task instructions given input instances and labels. \cite{singh2023explaining} introduces interpretable auto-prompting to generate natural-language explanations for data patterns. All these methods share the common goal of generating instructions for a given task from a few examples and may suffer from a lack of task diversity.  
To address the challenge of generating instructions corresponding to diverse tasks, a recent study by \cite{wang2023selfinstruct} proposed a semi-automated framework for task-agnostic instruction generation. Their framework begins with initial instructions and employs bootstrapping methods for expansion. Following a no-training paradigm, their framework encompasses various stages and relies on the large-parameter language model GPT-3.5 \cite{gpt35turbo} for inference. Notably, most of these methods rely on a non-open source commercial product, large in size, expensive, and subject to rate limits.

In this paper, we also focus on generating instructions along with the corresponding input-output pairs in a task-agnostic manner with negligible human intervention, effort, and cost. We mainly investigate a few unexplored questions. 
Specifically, we ask 1) How does the capability of low-parameter size language models like LLaMA 2 (7B, 13B) \cite{touvron2023llama} and Mistral 7B \cite{jiang2023mistral} compare to GPT-3.5 in generating instructions in a task-agnostic manner?
2) What is the performance analysis of the above models across varying numbers of instructions?
3) How effective is a training algorithm, such as Reinforcement Learning from Automated Feedback (RLAF), in the instruction-generation pipeline utilizing small parameter models like LLaMA 2-7B, LLaMa 2-13B, and Mistral 7B?


 

To investigate the aforementioned questions, we introduce, a semi-automated framework designed to generate high-quality instruction-input-output triplets for new tasks using Reinforcement Learning From Automated Feedback. 
Initially, our framework utilizes a small seed set of manually written tasks to generate instructions for defining new tasks (similar to 
Subsequently, reinforcement learning with automated feedback is employed to improve the quality of input-output pairs generated by passing instructions to the LLM. In the final stage, the model trained in the previous stage is utilized to construct the Instruction Fine Tuning dataset, serving as the basis for refining the base model 
through supervised fine-tuning.

To assess the effectiveness of our framework , we utilize LLaMa 2-7B, LLaMa 2-13B and Mistral-7B models to generate 15,000 instructions each, complete with input and outputs. Analysis of the resulting data reveals a diverse array of new tasks. Subsequently, we fine-tune the LLaMA 2-7B, LLaMA 2-13B, and Mistral 7B models using the generated instructions and evaluate their performance on the \textsc{Super-NI} dataset \cite{wang2022supernaturalinstructions}. Furthermore, we compare our framework with the self-instruct framework \cite{wang2023selfinstruct}, which utilizes LLaMA 2-7B, LLaMA 2-13B, and Mistral 7B models. Our findings demonstrate a significant task improvement of 64.39\%, 66.39\%, and 63.51\% across a wide range of tasks using LLaMA 2-7B, LLaMA 2-13B, and Mistral 7B as base models, respectively. Human evaluation of user-oriented tasks also further highlights the increment in performance. We also experiment with various dataset sizes proving that the method's effectiveness increases with the number of instructions. 

In summary, our contributions are:
\begin{itemize}
    \item  We empirically analyze the effectiveness of LLaMA 2-7B, LLaMA 2-13B, and Mistral 7B for task-agnostic instruction generation utilizing the existing framework.

    \item We introduce \ourmethod, a method for generating instructions with minimal human-labeled data, and LLaMa 2-7B, LLaMA 2-13B, Mistral 7B. We demonstrate its
effectiveness via extensive instruction-tuning experiments.

\item  We release a large synthetic
dataset of 45K instructions generated
by REFINE-AF using different LLM backbones.

\end{itemize}

\section{Methodolody}

\begin{figure*}[h!]
    \centering
    \includegraphics[width=0.8\textwidth]{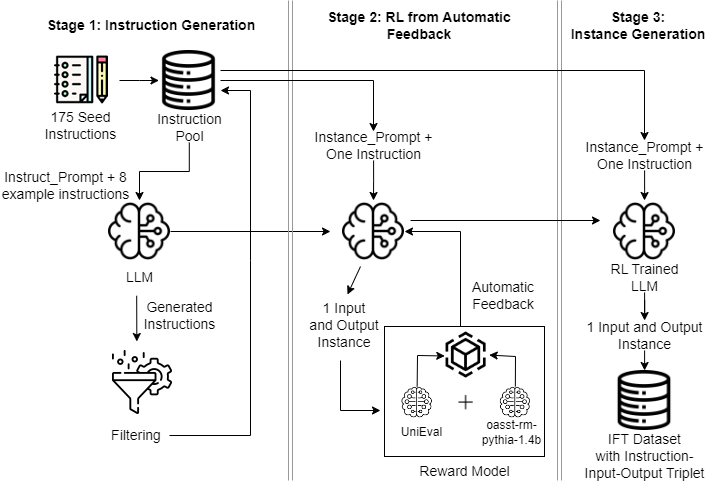}
    \caption{Schematic diagram of the stages in \ourmethod\ pipeline.}
    \label{fig:arch}
\end{figure*}

The task of annotating extensive instructional data poses challenges for human annotators, as it demands considerable time and effort, inventive thinking to devise fresh tasks and specialized knowledge to craft solutions for each task. We propose as a solution to this problem - \ourmethod, which generates synthetic instructional data from a small seed of human-written instructions by bootstrapping instructions using LLM inference, followed by training the LLM to align it to automated preferences in generated (input, output) pairs. Here, we elaborate our pipeline for our proposed \ourmethod\ in 3 stages (see Figure \ref{fig:arch}) - (1) Instruction Generation (2) Using RL from Automated Feedback to generate input-output pairs (3) Instance Generation. The final output of \ourmethod\ is a dataset of (instruction, input, output) triplets, which could later be used for updating an LLM via Instruction Fine-Tuning (IFT).

\subsection{Notation}

Let us define the notation utilized in this paper. $\pi^{0}$ represents the foundational LLM responsible for generating instructions. $\pi^{RL}_{\phi}$ denotes the LLM undergoing training via the Reinforcement Learning (RL) pipeline, where $\phi$ signifies the parameter set within the LLM. $x$ stands for the input provided to the LLM, comprising the prompt and the instruction. Correspondingly, $y$ represents the response generated by the LLM, encompassing the input-output instances. Furthermore, $r(x,y)$ is employed to signify the reward obtained from the reward model utilized for guiding the LLM. Various terms utilized in the reward function are elaborated in Table \ref{tab:indicators}.

\subsection{Stage 1:Instruction Generation}

\ourmethod\ generates additional instructions by iteratively building upon a limited set of initial human-written instructions. The initial pool of instructions is initiated using 175 seed instructions like \cite{wang2023selfinstruct}. At every step, we randomly select 8 instructions from the pool to serve as in-context examples. 6 of the 8 instructions are written by humans, while the remaining 2 are generated by the LLM in the preceding steps to ensure diversity. To promote diversity, a new instruction is introduced into the instruction pool only if its ROUGE-L similarity score with any existing instruction is below 0.7 maintaining the approach followed by \cite{wang2023selfinstruct}.  Additionally, we eliminate instructions containing certain keywords (such as image, picture, graph) that are typically not processable by LLMs. The prompts used for the same are given in Appendix \ref{app:instruction_gen_prompt}.

\subsection{Stage 2:Using RL from Automated Feedback to generate input-output pairs}

\begin{table}[t]
\scalebox{0.45}{
\centering
\resizebox{1.0\textwidth}{!}{
\begingroup
\setlength{\tabcolsep}{6pt} 
\renewcommand{\arraystretch}{1.5} 
    \begin{tabular}{ccl}
        \toprule
        \textbf{Indicator} & \textbf{Notation} &\textbf{Explanation}\\
        \midrule   
        {Reward score} & $Rew$ & \makecell[l]{The \texttt{oasst-rm-pythia-1.4b} reward \\ model inference score \cite{köpf2023openassistant}} \\\specialrule{0em}{1pt}{1pt}
        {Unieval-naturalness} & $Nat$ & \makecell[l]{Whether a response is natural, provided \\ by the UniEval \cite{zhong2022unified} \\ dialogue model.} \\\specialrule{0em}{2pt}{2pt}
        {Unieval-coherence} & $Coh$ & \makecell[l]{Whether this response is as a valid \\ continuation of the previous conversation, \\ provided by the UniEval \cite{zhong2022unified}\\ dialogue model.} \\\specialrule{0em}{2pt}{2pt}
        {Unieval-understandability} & $Und$ & \makecell[l]{Whether the response is understandable, \\ provided by the UniEval \cite{zhong2022unified}\\ dialogue model.} \\\specialrule{0em}{2pt}{2pt}
        \bottomrule
    \end{tabular}

\endgroup
 }
}
\caption{Summary of indicators for instruction quality. Each data sample is viewed as a pair of instruction and response (i.e., input and output) of LLM.}
\label{tab:indicators}
\end{table}

RL from Human Feedback (RLHF) \cite{christiano2017deep,stiennon2020learning} leverages human preferences as a form of reward signal for refining LMs and aligning them to human judgements. RLHF essentially comprises a reward model that receives a text sequence as input and outputs a single scalar reward, intended to quantitatively reflect human preference. This scalar reward serves as feedback for the LLM, facilitating its parameter updates using a policy-gradient RL algorithm such as Proximal Policy Optimization (PPO) \cite{ppo}.

In \ourmethod, in order to keep human effort to a minimum, we replace human feedback with automated feedback. The quality of the instruction data could be viewed as its ability to efficiently steer language models in learning to generate responses in a particular manner. This can be estimated by various indicators as described in Table \ref{tab:indicators}.

Adapting from an earlier work by \cite{cao2023instruction}, the reward score for any instruction, input, output triplet is calculated as: 

\begin{equation}
\begin{aligned}
\label{eq:score}
r(x,y) & = 0.0078 \times Rew(x,y)  \\
& \quad  - 0.4421 \times Und(x,y) \\
& \quad + 0.3212 \times Nat(x,y)  \\
& \quad  + 0.1520 \times Coh(x,y) - 0.0274
\end{aligned}
\end{equation}

Table \ref{tab:indicators} elaborates each term used in Equation \ref{eq:score}. This score acts as a scalar notion of preferability for a particular instruction-input-output triplet. It is directly proportional to the \texttt{oasst-rm-pythia-1.4b} model reward \cite{köpf2023openassistant}, the naturalness and the coherence of the triplet and inversely proportional to the understandability which represents the complexity in the sentence. These metrics are obtained from the UniEval \cite{zhong2022unified} dialogue model.

In the training phase, the LLM is engaged through a specifically designed prompt containing examples of Instruction-Input-Output Instances and instructions to generate such instances given an instruction. The prompt is provided in the Appendix \ref{app:instance_gen_prompt}. The model output is then passed to the reward model described above and an automatic score is generated which serves as the feedback to the LLM. 

We maximise the following objective function in RL training using the PPO algorithm considering the Kullback-Leibler (KL) divergence factor: 

\begin{align}
R(x,y) &= r(x,y) - \beta \log \left(\pi_{\phi}^{\mathrm{RL}}(y \mid x) / \pi^{\mathrm{0}}(y \mid x)\right)
\end{align} 
\text{where:} \nonumber
\begin{itemize}
    \item The second term is the KL divergence between the current policy and the reference model as denoted by \cite{ouyang2022training},
    \item $\beta$ is a scaling factor ($\beta > 0$).
\end{itemize} \nonumber

\subsection{Stage 3:Instance Generation}

After training the LLM using RL with automated feedback in the previous stage, we use the same prompt used while training followed by 1 instruction from the instruction pool at a time. This generates a (input, output) pair corresponding to each input instruction. At the end of this stage, we get an Instruction Fine Tuning (IFT) dataset of (instruction, input, output) triplets, as desired in a semi-automated fashion.

\textbf{Instruction Finetuning:} Following the instance generation phase, the generated IFT dataset serves as the foundation for refining the model through Supervised Finetuning (SFT), a technique prevalently adopted for instruction finetuning.

\section{Experiments}
In this section, we conduct experiments to measure and compare the performance of \ourmethod\ with other instruction dataset generation pipelines with various base models.
\begin{figure}[H]
    \centering
    \includegraphics[width= 0.9\linewidth]{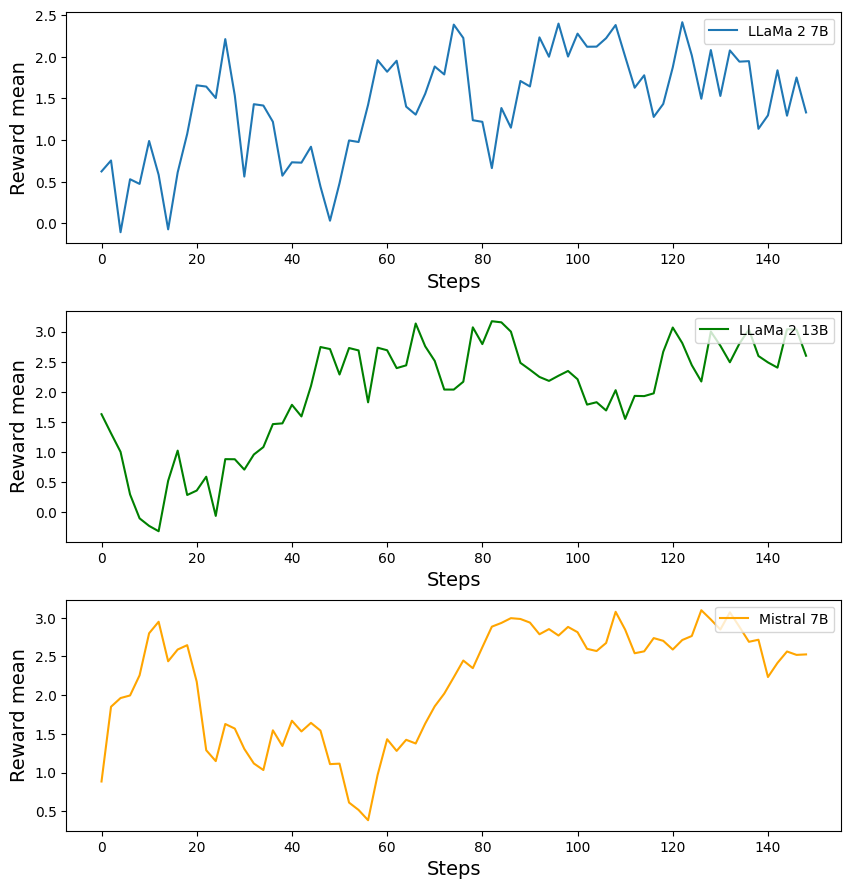}
    \caption{Moving average of the model rewards over the training steps for LLaMa 13B, LLaMa 13B and Mistral 7B.}
    \label{fig:reward}
\end{figure}

\subsection{Experimental Setup}

\textbf{Seed Data:} We use the same set of 175 seed examples utilized by Self-Instruct \cite{wang2023selfinstruct} as human-annotated seed data for bootstrapping in the initial stage.\\
\noindent \textbf{Base Models and Hyper-parameter setting. }We use the following pre-trained models of different sizes - LLaMa 2 model \cite{touvron2023llama} with 7B, 13B parameters, and Mistral \cite{jiang2023mistral} with 7B parameters as the base models for generating the instruction dataset.
We use the PPO Trainer from the  Transformer Reinforcement Learning (TRL) library\footnote{\url{https://huggingface.co/docs/trl/en/index}}. The model is loaded in 4-bit mode and trained with LoRA (Low-Rank Adaptation) \cite{lora} for 200 steps with a batch size of 4, learning rate 2e-5, and 4 gradient accumulation steps. For the supervised fine-tuning step (using the generated instruction dataset), we train the model for 3 epochs, with a learning rate of 2e-5, cosine scheduler with a warmup ratio of 0.3, batch size of 2 with 4 gradient accumulation steps using the HuggingFace Trainer API\footnote{\url{https://huggingface.co/docs/transformers/main/en/trainer}}. All models were trained using a single A100 GPU. The initial phase of our process requires approximately 20 days to generate 15K instructions, and none of the subsequent phases require more than 120 hours. We filled out the consent form to use the LLaMa 2 model.\\
\noindent \textbf{Baselines.} To evaluate the effectiveness of \ourmethod, we compare \ourmethod\ with SELF INSTRUCT framework applied on previously mentioned LLM backbones (LLaMa 2 7B, LLaMa 2 13B \cite{touvron2023llama}, Mistral 7B \cite{jiang2023mistral}) of different sizes as baselines. \\
\textbf{RL Training: } Training Language Model Models (LLMs) with Reinforcement Learning (RL) presents its challenges, as noted by Beeching et al. (2023) \cite{beeching2023stackllama}, often accompanied by instability. To gauge the efficacy of the training process, we monitor the model's mean reward across training steps. To ensure clarity, we employ a moving average with a span of 30 steps. The Spearman rank correlation between reward and steps, illustrated in Figure \ref{fig:reward}, stands at 0.553, 0.649, and 0.558 for LLaMa 7B, LLaMa 13B, and Mistral 7B, respectively. A notably positive Spearman rank correlation suggests a progression where the system gains proficiency over time, with a robust inclination for reward enhancement as training advances.

\section{Data Quality of Generated Instruction Dataset}
In this section, we discuss the generated instruction dataset using \ourmethod. The statistical details of the generated data are outlined in Appendix \ref{sec:st}.
\subsection{Analysis of the generated instances}
In this section, we'll explore the quality and diversity of the generated instances.

\noindent \textbf{Diversity:} To evaluate the diversity of the generated instructions, we follow the approach in \cite{wang2023selfinstruct} and utilize the Berkeley Neural Parser to analyze the instructions and subsequently identify the verb along with its immediate noun object. 
The instructions generated by \ourmethod\ using LLama 2-7B, LLama 2-13B, and Mistral 7B comprise a total of 828, 790 and 467 unique noun-verb pairs, respectively. Figures \ref{fig:nounverb_7b}, \ref{fig:nounverb_13b}, and \ref{fig:nounverb_mistral} in Appendix display the 20 most frequent root verbs and their top 4 associated noun objects for the models LLama 2-7B, LLama 2-13B, and Mistral 7B, respectively.

\begin{table*}[!ht]
\centering
\begin{tabular}{lllll}
\hline
\multicolumn{1}{c}{Model}    & \multicolumn{1}{c}{Instruction Count} & \multicolumn{1}{c}{SELF INSTRUCT} & \multicolumn{1}{c}{REFINE-AF} & \% Task Better \\ \hline
\multirow{3}{*}{LLaMa 2 7B}  & 5,000                                 & 5.8012                            & \textbf{5.9613}               & 66.66\%        \\
                             & 10,000                                & 5.9841                            & \textbf{6.0398}               & 53.79\%        \\
                             & 15,000                                & 6.0414                            & \textbf{6.1636}               & 64.39\%        \\ \hline
\multirow{3}{*}{LLaMa 2 13B} & 5,000                                 & \textbf{6.5349}                            & 6.4488                        & 44.54\%        \\
                             & 10,000                                & 6.5269                            & \textbf{6.5381}               & 52.94\%        \\
                             & 15,000                                & 6.4446                            & \textbf{6.6133}               & 66.39\%        \\ \hline
\multirow{3}{*}{Mistral 7B}  & 5,000                                 & 5.7615                            & \textbf{5.8632}               & 64.29\%        \\
                             & 10,000                                & 5.8454                            & \textbf{5.9712}               & 60.34\%        \\
                             & 15,000                                & 5.9986                            & \textbf{6.1348}               & 63.51\%        \\ \hline
\end{tabular}
\caption{Comparative results (Average ROUGE-L Scores) of \ourmethod\ with Self Instruct method for different base models on 119 tasks in \textsc{Super-NI}.}
\label{tab:res}
\end{table*}

\begin{table*}[!t]
\centering
\begin{tabular}{l|rr|rr|rr}
\hline
\multicolumn{1}{c|}{\multirow{2}{*}{Task Category}} & \multicolumn{2}{c|}{LLaMa 2 7B}                                    & \multicolumn{2}{c|}{LLaMa 2 13B}                                   & \multicolumn{2}{c}{Mistral 7B}                                    \\ \cline{2-7} 
\multicolumn{1}{c|}{}                               & \multicolumn{1}{c}{SELF INSTRUCT} & \multicolumn{1}{c|}{REFINE-AF} & \multicolumn{1}{c}{SELF INSTRUCT} & \multicolumn{1}{c|}{REFINE-AF} & \multicolumn{1}{c}{SELF INSTRUCT} & \multicolumn{1}{c}{REFINE-AF} \\ \hline
title\_generation                                   & \textbf{8.0902}                   & 8.017                          & 8.4642                            & \underline{\textbf{8.7679}}                & \textbf{7.8912}                   & 7.7654                        \\
coreference\_resolution                             & 2.1905                            & \underline{\textbf{2.3077}}                & \textbf{2.3639}                   & 2.1891                         & \textbf{2.2134}                   & 2.1543                        \\
textual\_entailment                                 & \textbf{1.7029}                   & 1.6419                         & \textbf{2.0271}                   & 2.0019                         & \textbf{1.7543}                   & 1.6312                        \\
question\_rewriting                                 & 20.2548                           & \underline{\textbf{20.6361}}               & 20.5859                           & \underline{\textbf{21.6007}}               & 20.0412                           & \underline{\textbf{20.3401}}              \\
cause\_effect\_classification                       & 5.1514                            & \underline{\textbf{5.3748}}                & \textbf{5.4321}                   & 5.3932                         & 5.0768                            & \underline{\textbf{5.2312}}               \\
dialogue\_act\_recognition                          & 2.0877                            & \textbf{2.1171}                & 2.4235                            & \textbf{2.4896}                & 2.0421                            & \textbf{2.1045}               \\
answerability\_classification                       & 1.5365                            & \underline{\textbf{1.6594}}                & 1.9846                            & \textbf{2.0515}                & 1.4571                            & \underline{\textbf{1.5762}}               \\
keyword\_tagging                                    & 3.0131                            & \underline{\textbf{3.1371}}                & 3.8338                            & \textbf{3.8830}                & 3.0021                            & \underline{\textbf{3.1127}}               \\
data\_to\_text                                      & 15.4563                           & \underline{\textbf{15.6081}}               & 15.2005                           & \underline{\textbf{15.5802}}               & 15.1112                           & \underline{\textbf{15.3456}}              \\
word\_analogy                                       & 1.2353                            & \textbf{1.2812}                & \textbf{1.3527}                   & 1.3294                         & \textbf{1.1872}                   & 1.1239                        \\
overlap\_extraction                                 & 5.4254                            & \underline{\textbf{5.6473}}                & 14.3674                           & \underline{\textbf{15.2493}}               & 6.1231                            & \underline{\textbf{7.2367}}               \\
grammar\_error\_correction                          & 31.5197                           & \textbf{31.586}                & 36.8425                           & \underline{\textbf{37.2230}}               & 30.1172                           & \underline{\textbf{31.2387}}              \\ \hline
\end{tabular}
\caption{Task Category Wise Comparison of the average ROUGE-L Scores between the pipelines. REFINE AF outperforms SELF INSTRUCT in the majority of the categories of \textsc{Super-NI} Dataset for all the base models. The underlined values denote significantly larger improvements (p-value < 0.05) using statistical significance test}
\label{tab:taskwiseres}
\end{table*}

\begin{figure}[H]
    \centering
    \includegraphics[width=0.8\linewidth]{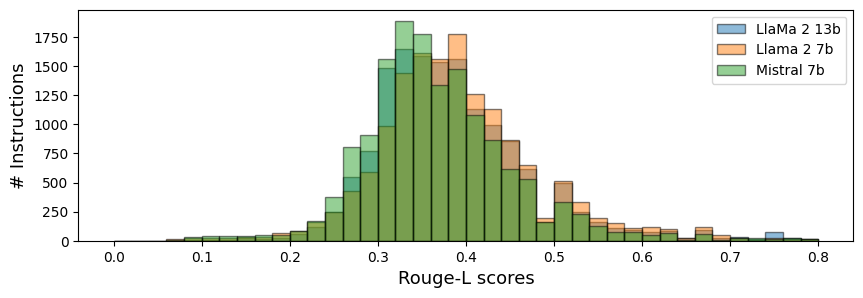}
    \caption{Distribution of Rouge-L scores between generated instructions and their most similar seed instruction. This highlights the difference in generated instructions from seed tasks.}
    \label{fig:rouge}
\end{figure}

We further investigated the disparity between the generated and seed instructions utilized to initiate the generation process. Specifically, for each generated instruction, we calculated its maximum ROUGE-L overlap with the pool of 175 seed instructions. We depict the distribution of these ROUGE-L scores in Figure \ref{fig:rouge}. The findings reveal a notable proportion of newly generated instructions by REFINE-AF using LLaMA 2-7B, LLaMA 2-13B, and Mistral 7B having minimal overlap with the seed instructions.

\begin{figure}[H]
    \centering
    \includegraphics[width=0.8\linewidth]{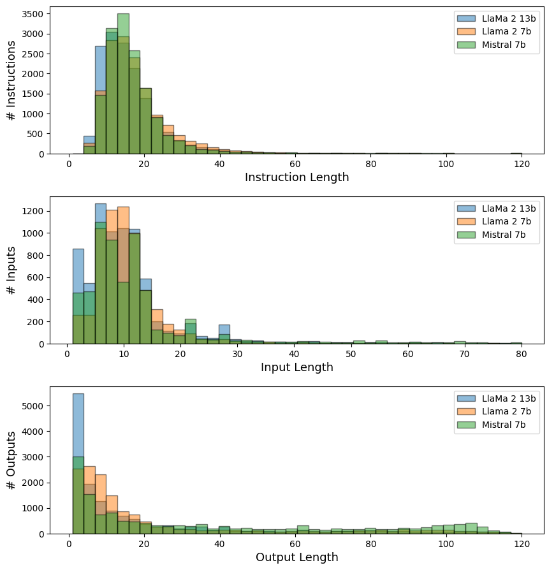}
    \caption{Length distributions of the instructions, inputs, and outputs generated by \ourmethod.}
    \label{fig:length}
\end{figure}

\begin{figure}[H]
    \centering
    \includegraphics[width=0.8\linewidth]{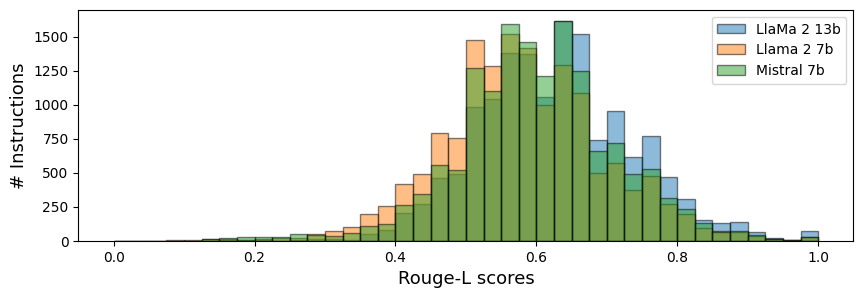}
    \caption{Similarity of instructions generated using \ourmethod\ wrt the GPT 3.5 instructions for different base models. This similarity has been calculated using Rouge-L scores.}
    \label{fig:gpt3similarity}
\end{figure}

Figure \ref{fig:length} describes the distribution of the length of the instructions, inputs, and outputs in the generated instances\footnote{from here onwards, an `instance' refers to an instruction-input-output triplet}. This further highlights the diversity of the generated instruction dataset. Most of these observations match the ones generated by instructions using the Self-instruct method on GPT-3.5 \cite{wang2023selfinstruct}, showing that even smaller language models are capable of generating good-quality instructions. 

We also analyzed instructions generated by LLaMA 2-7B, LLaMA 2-13B, and Mistral 7B with GPT-3.5. Specifically, we examined 15,000 instructions generated by \ourmethod for each of the models LLama 2-7B, LLama 2-13B, and Mistral 7B. For GPT-3.5, we utilized the dataset introduced by \cite{wang2023selfinstruct}. Utilizing the Rouge-L score, we determined the most similar instruction for each model-generated instruction. The distribution of similarity scores is depicted in Figure \ref{fig:gpt3similarity}. Notably, many instructions closely resembled those generated by the GPT-3.5 model, with a mean score of approximately 0.62. This observation underscores the capability of even smaller LLMs to produce high-quality instructions. 

\textbf{Quality:} In this section, we evaluate the quality of our generated instructions. We randomly select 100 instructions from each set of model-generated instructions for this analysis, sampling one instance per instruction. Following the methodology outlined in \cite{wang2023selfinstruct}, we ask an expert annotator (including one of the authors of this work) to evaluate whether each instance is correct in terms of the instruction, the instance input, and the instance output. The data quality assessment presented in Table \ref{table:quality} within the Appendix indicates that the majority of the generated instructions are accurate. While there are some instances of noise in both the input and output associated with these instructions, the overall findings are promising. Despite these minor discrepancies, all the instructions remain valid and hold potential utility for model training purposes. 

\section{Experimental Results}

\subsection{Zero-Shot Generalization on the \textsc{Super-NI} benchmark}

Initially, we assess the model's proficiency in adhering to instructions across conventional NLP tasks in a zero-shot manner. This evaluation is conducted on the evaluation set of the \textsc{Super-NI} dataset \cite{wang2022supernaturalinstructions}, using the methodology adopted by \cite{wang2023selfinstruct}. The \textsc{Super-NI} dataset comprises 119 tasks across 12 task categories, each comprising 100 instances. The models are not provided with any in-context examples to aid comprehension; they are solely prompted with task definitions.

\textbf{Results:} The experimental results are presented in Table \ref{tab:res}. Our proposed method consistently surpasses the baseline SELF-INSTRUCT approach across various instruction set sizes. Evaluation in the \textsc{SuperNI} dataset employs both exact match and Rouge-L metrics. Since \cite{wang2023selfinstruct} solely reported Rouge-L scores, we adhere to presenting Rouge-L metrics.
While for classification tasks, similarity scores with labels are typically utilized, this practice was not adopted in the \textsc{SuperNI} dataset, hence we rely on the Rouge-L metrics exclusively. 

Notably, it exhibits superior performance across different tasks, surpassing the baseline in the majority of them. Further insights into the performance based on task categories are presented in Table \ref{tab:taskwiseres}. Remarkably, our pipeline demonstrates enhanced performance compared to the baseline in 10 out of 12 task categories, underscoring the method's robustness across diverse task domains. Moreover, the performance consistency across different models suggests the scalability of our approach, indicating its potential applicability to larger models to enhance the quality of instruction generation.

\subsection{Generalisation of User-oriented Instructions on Novel Tasks}

The \textsc{Super-NI} covers an extensive set of existing NLP tasks, but these were proposed for research purposes and are skewed towards classification. To better evaluate the instruction following capabilities of the models we also experiment on user-oriented instructions created by \cite{wang2023selfinstruct}. These contain 252 instructions with 1 instance per instructions which cover diverse and unfamiliar instructions. 

\textbf{Human Evaluation:} The instructions utilized in this experiment are designed to be more generic and user-oriented. Given the diverse nature of tasks, different expertise is required, rendering them unsuitable for evaluation via automatic metrics or standard crowd workers. Consequently, the authors of this paper conducted manual evaluations, referencing the golden outputs provided with these instructions. Evaluators rated the outputs into one of several categories based on how effectively the model achieved the task. Following the framework introduced by \cite{wang2023selfinstruct}, a four-level rating system was adopted:
\begin{itemize}
\item RATING-A: The response is valid and satisfactory.
\item RATING-B: The response is acceptable but contains minor errors or imperfections.
\item RATING-C: The response is relevant and addresses the instruction, but significant errors are present.
\item RATING-D: The response is irrelevant or completely invalid.
\end{itemize}

\begin{figure}[t]
    \centering
    \includegraphics[width=0.99\linewidth]{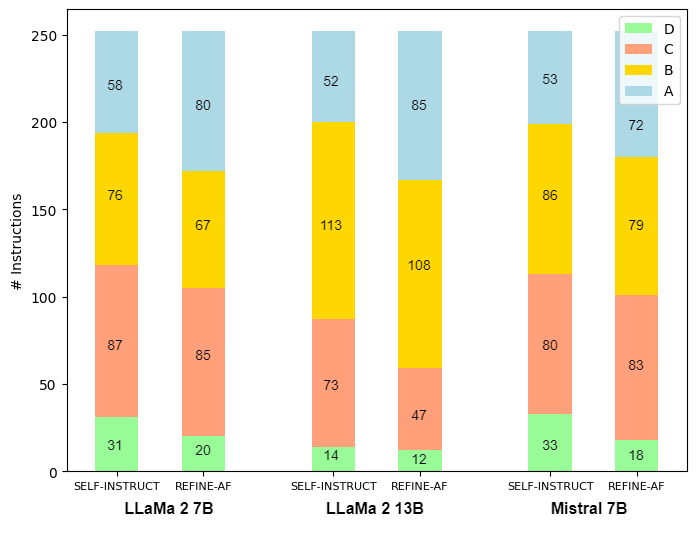}
    \caption{Human Evaluation results using LLaMA 7B, LLaMA 2-13B, and Mistral 7B models trained on 15k instructions utilizing Self Instruct and our pipeline. }
    \label{fig:humaneval}
\end{figure}

\textbf{Results: } Figure \ref{fig:humaneval} illustrates the outcomes of the human evaluation. Annotators evaluated the responses from each model separately, without knowledge of the method employed, to mitigate bias. We can observe that \ourmethod\ performs better than SELF-INSTRUCT as it can get valid and satisfactory answers to more instructions. Also, the count of irrelevant responses is smaller. Thus the RL feedback guides the model to generate better responses to instructions.

\begin{figure}[]
    \centering
    \includegraphics[width=0.9\linewidth]{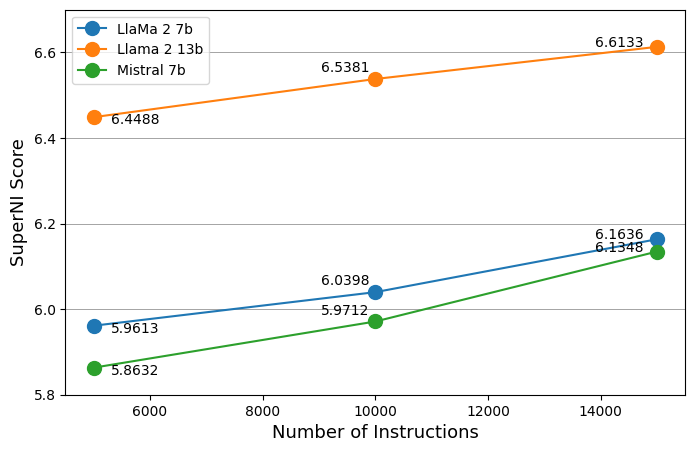}
    \caption{Effect of data size on the performance of the model. Shown using the scores on \textsc{Super-NI} benchmark}
    \label{fig:datasize}
\end{figure}

\subsection{Effect of Data Size and Quality}
\ourmethod\ provides a way to generate a large number of good-quality instructions. But does increasing the number of instructions improve the performance? We evaluate the following in this section. We can observe in Table \ref{tab:res} that increasing the number of instructions gradually increases the score on the \textsc{Super-NI} benchmark. A more descriptive plot of the same can be found in Figure \ref{fig:datasize}. 
Thus increasing the number of instructions proves to be beneficial to the performance. 

\section{Related Work}
\textbf{Instruction tuning for LLMs:} 
Early investigations \cite{wei2022finetuned,mishra-etal-2022-cross,sanh2022multitask} into instruction fine-tuning primarily focused on natural language processing (NLP) tasks. They revealed that fine-tuning with NLP datasets structured as instruction-output pairs enhances cross-task generalization. Additionally, a few studies \cite{wang2022supernaturalinstructions,chung2022scaling} have demonstrated a direct correlation between the extent and diversity of the "instructional" data and the model's capacity to generalize to unfamiliar tasks. Works such as \cite{ouyang2022training} and \cite{wang2023selfinstruct} aim to construct general-purpose Language Models capable of handling a broader array of task instructions, aligning with our objectives. However, the creation of diverse datasets and the utilization of large-parameter-based, non-open-source models pose significant challenges. 
\\ 
\textbf{Instruction Generation:} A significant hurdle in enabling Large Language Models (LLMs) to follow instructions is the collection of demonstration samples for fine-tuning. Current high-accuracy LLMs heavily rely on human annotators, and most instruction datasets are not open source. Moreover, these human annotation approaches are time-consuming, costly, and require expertise across various domains. To mitigate these challenges, several methods have been proposed for automated instruction generation. For instance, \cite{köksal2024longform} utilize human-written text as a natural response and generate corresponding instructions based on the response. \cite{li2024selfalignment} employ a semi-automated method using a back-translation approach to generate instructions from human-written text data.  

Several studies \cite{zhou2023large,ye2023guess,honovich2022instruction,singh2023explaining} have investigated the utilization of Large Language Models (LLMs) for instruction generation. Employing unnatural instruction prompts, GPT-3 generates additional instructions when provided with a few initial seed instructions within a contextual setting. \cite{wang2023selfinstruct} proposed a semi-automated framework for task-agnostic instruction generation. Their framework begins with initial instructions and employs bootstrapping methods for expansion. Following a no-training paradigm, their framework encompasses various stages and relies on the large-parameter language model GPT-3.5 \cite{gpt35turbo} for inference. In contrast, our method utilizes small-parameter-sized open-source LLMs and employs a reinforcement learning algorithm through human feedback to generate instructions directly from the model itself. \\
\textbf{Model Self-training:} A conventional self-training framework, as demonstrated by \cite{he2020revisiting,xie2020selftraining,du-etal-2021-self,amini2023selftraining,huang2022large,zhou2022prompt}, entails using trained models to label unlabeled data and then utilizing this newly labeled data to enhance the model's performance. 
\section{Conclusions and Future Work}
 In this study, we have conducted an empirical analysis to assess the effectiveness of three open-source, low-parameter models—LLaMA 2-7B, LLaMA 2-13B, and Mistral 7B—in comparison to the larger model GPT-3.5 for task-agnostic instruction generation within an established framework, SELF-INSTRUCT. Additionally, we have introduced \ourmethod, a novel approach for generating instructions in a task-agnostic manner with minimal human-labeled data, leveraging the capabilities of LLaMa 2-7B, LLaMA 2-13B, and Mistral 7B. Through extensive experimentation in instruction-tuning, we have demonstrated the effectiveness of our approach. Furthermore, we have made a significant contribution by releasing a large synthetic dataset comprising 45K instructions generated by REFINE-AF using different LLM backbones, thus facilitating further research in this domain.

 \section{Limitations}

 The primary limitation of our instruction generation framework, which comprises three parts, lies in the initial stage responsible for instruction generation. This phase tends to consume significant time when generating instructions directly from the model. In future iterations, we aim to integrate more efficient methods to create instructions within shorter timeframes. Our model has been evaluated on various NLP tasks, and this work can extend to include multimodal scenarios.

 \section{Appendix}
\label{sec:appendix}

\begin{table*}[]
\centering
\begin{tabular}{llll}
\hline
\multicolumn{1}{c}{Quality Review Question}   & \multicolumn{1}{c}{LlaMa2 7B} & \multicolumn{1}{c}{LlaMa2 13B} & \multicolumn{1}{c}{Mistral 7B} \\ \hline
Does the instruction \\
describe a valid task?             & 90\%                      & 94\%                        &      95\%                   \\

Is the input appropriate\\
for the instruction? & 81\%                         & 83\%                         &       76\%                   \\

Is the output a correct and acceptable \\
response to the instruction and input?            & 58\%                        & 65\%                        &            64\%           \\ \hline
\end{tabular}
\caption{Evaluation of the data quality for the instruction, input, and output of the generated data by REFINE-AF with different LLM backbones}
\label{table:quality}
\end{table*}

\subsection{Statistics of Instructions Generated}\label{sec:st}

Table \ref{table:stats} presents the fundamental statistics about the generated data. 15,000 instructions were generated in Stage 1, resulting in an instruction dataset of around 15,000 instances corresponding to each instruction post-filtering. 

\begin{table*}
\centering
\begin{tabular}{llll}
\hline
\multicolumn{1}{c}{Statistics}   & \multicolumn{1}{c}{LlaMa2 7B} & \multicolumn{1}{c}{LlaMa2 13B} & \multicolumn{1}{c}{Mistral 7B} \\ \hline
\# of instructions               & 15,000                        & 15,000                         & 15,000                         \\
\# of instances                  & 14,998                        & 17,524                         & 14,953                         \\
\# of instances with empty input & 8,564                         & 7,504                          & 8,749                          \\
average instruction length       & 17.77                         & 15.63                          & 16.82                          \\
average non-empty input length   & 10.34                         & 10.50                          & 9.64                           \\
average output length            & 22.97                         & 18.43                          & 24.56                          \\ \hline
\end{tabular}
\caption{Statistics of the instructions generated by \ourmethod\ with different LLM backbones}
\label{table:stats}
\end{table*}

\subsection{Diversity of generated instructions}
\begin{figure*}
    \centering
    \begin{subfigure}[b]{0.49\textwidth}
        \includegraphics[width=\textwidth]{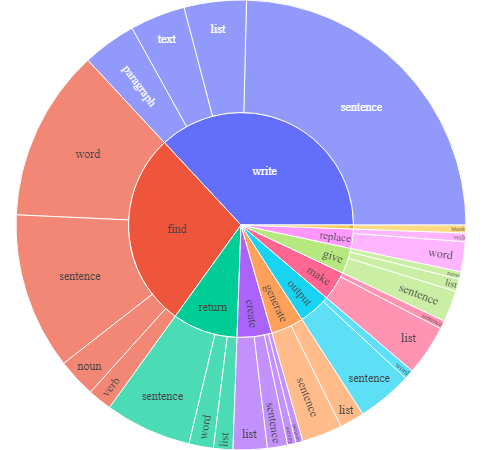}
        \caption{Llama 2 7B}
        \label{fig:nounverb_7b}
    \end{subfigure}
    \hfill
    \begin{subfigure}[b]{0.49\textwidth}
        \includegraphics[width=\textwidth]{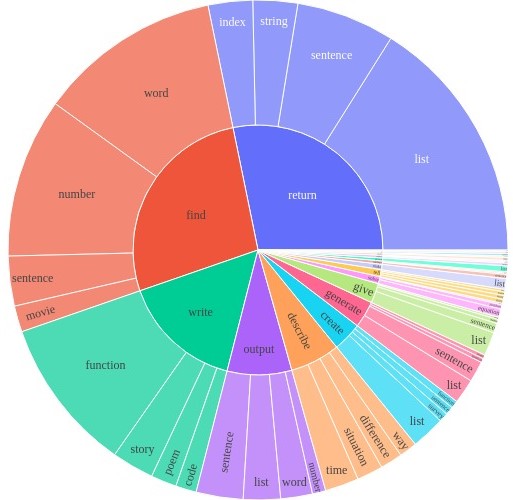}
        \caption{Llama 2 13B}
        \label{fig:nounverb_13b}
    \end{subfigure}
    
    \begin{subfigure}[b]{0.49\textwidth}
        \includegraphics[width=\textwidth]{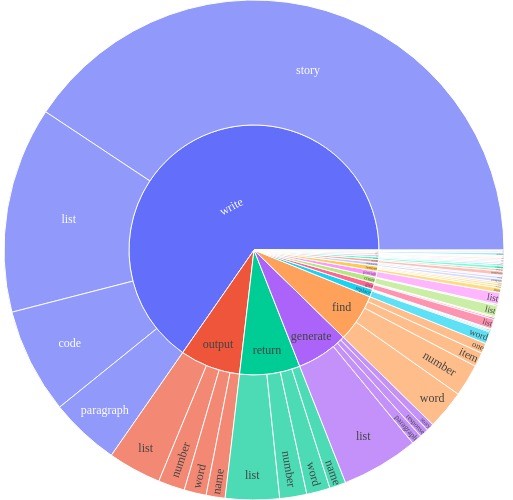}
        \caption{Mistral 7B}
        \label{fig:nounverb_mistral}
    \end{subfigure}
    
    \caption{The top 20 most common root verbs (inner circle) and their top 4 most common direct nouns (outside circle) in the generated instructions that contain such verb-noun structure for different models.}
    \label{fig:nounverb_combined}
\end{figure*}

As discussed in the paper, we have observed the root verbs with their main nouns to evaluate the diversity of the generated instructions. Figure \ref{fig:nounverb_combined} displays the plot highlighting the top 20 most common verbs in the inner circle followed by the top 4 most common direct nouns in the outer circle for the generated instructions for different models.

\subsection{Prompt Template for Instruction Generation}
\label{app:instruction_gen_prompt}

We have modified the prompt used by Self Instruct to be more comprehensive and give detailed instructions regarding instruction generation. This prompt has been used to train Stanford's Alpaca model \cite{alpaca} which was released after the Self Instruct paper and was based on the Self Instruct data pipeline. For a fair comparison of our pipeline and Self Instruct, we use the same template for instruction generation in both pipelines. The updated prompt is shown in Table \ref{tab:instruction_generation_template}.

\begin{table*}[ht]
    \centering
    \small
    \noindent\fbox{%
    \begin{minipage}{\dimexpr\linewidth\fboxsep-2\fboxrule} 
\tt 
You are asked to come up with a set of diverse task instructions. These task instructions will be given to a LLM and we will evaluate the LLM for completing the instructions. \\
Here are the requirements: \\
1. Try not to repeat the verb for each instruction to maximize diversity. \\
2. The language used for the instruction also should be diverse. For example, you should combine questions with imperative instructions. \\
3. The type of instructions should be diverse.  \\
4. The list should include diverse types of tasks like open-ended generation, classification, editing, etc. \\
5. A language model should be able to complete the instruction. For example, do not ask the assistant to create any visual or audio output. For another example, do not ask the assistant to wake you up at 5 pm or set a reminder because it cannot perform any action. \\
6. The instructions should be in English.\\
7. The instructions should be 1 to 2 sentences long. Either an imperative sentence or a question is permitted. \\
\\
Task 1: \{instruction for existing task 1\} \\
Task 2: \{instruction for existing task 2\} \\
Task 3: \{instruction for existing task 3\} \\
Task 4: \{instruction for existing task 4\} \\
Task 5: \{instruction for existing task 5\} \\
Task 6: \{instruction for existing task 6\} \\
Task 7: \{instruction for existing task 7\} \\
Task 8: \{instruction for existing task 8\} \\
Task 9:
    \end{minipage}
}
    \caption{Prompt used for generating new instructions. 8 existing instructions are randomly sampled from the task pool for in-context demonstration. The model can generate instructions for new tasks until it stops its generation, reaches its length limit or generates up to 8 more tasks.}
    \label{tab:instruction_generation_template}
\end{table*}

\subsection{Prompt Template for Instance Generation}
\label{app:instance_gen_prompt}

For instance generation, we use the same prompt as used by Self Instruct. It consists of a set of in-context demonstrations to demonstrate how input-output examples should look like. It is shown in Table \ref{tab:instance_generation_template}.

\begin{table*}[ht]
    \centering
    \small
    \noindent\fbox{%
    \begin{minipage}{\dimexpr\linewidth\fboxsep-2\fboxrule} 
\tt 
Come up with examples for the following tasks. Try to generate multiple examples when possible. If the task doesn't require additional input, you can generate the output directly.\\
\\
Task: Which exercises are best for reducing belly fat at home?\\
Output:\\
- Lying Leg Raises\\
- Leg In And Out\\
- Plank\\
- Side Plank\\
- Sit-ups\\
\\
Task: Extract all the country names in the paragraph, list them separated by commas.\\
Example 1\\
Paragraph: Dr. No is the sixth novel by the English author Ian Fleming to feature his British Secret Service agent James Bond. Written at Fleming's Goldeneye estate in Jamaica, it was first published in the United Kingdom by Jonathan Cape in 1958. In the novel Bond looks into the disappearance in Jamaica of two fellow MI6 operatives who had been investigating Doctor No. Bond travels to No's Caribbean island and meets Honeychile Rider, who is there to collect shells. They are captured and taken to a luxurious facility carved into a mountain. The character of Doctor No, the son of a German missionary and a Chinese woman, was influenced by Sax Rohmer's Fu Manchu stories. Dr. No was the first of Fleming's novels to face widespread negative reviews in Britain, but it was received more favourably in the United States.\\
Output: English, British, Jamaica, the United Kingdom, German, Chinese, Britain, the United States.\\
\\
Task: Converting 85 F to Celsius.\\
Output: 85°F = 29.44°C\\
\\
Task: Sort the given list ascendingly. \\
Example 1\\
List: [10, 92, 2, 5, -4, 92, 5, 101]\\
Output: [-4, 2, 5, 5, 10, 92, 92, 101]\\
Example 2\\
Input 2 - List: [9.99, 10, -5, -1000, 5e6, 999]\\
Output: [-1000, -5, 9.99, 10, 999, 5e6]\\
\\
Task: Suggest a better and more professional rephrasing of the following sentence.\\
Example 1\\
Sentence: This house is surprisingly not constructed very well, and you probably need more money to fix it after you buy it. If you ask me, I would suggest you to consider other candidates.\\
Output: This house does not seem to be constructed well, so you may need to spend more money to fix it after you purchase it. I would suggest that you look at other properties.\\
Example 2\\
Sentence: Just so you know, we did an experiment last week and found really surprising results - language model can improve itself!\\
Output: Our experiments last week demonstrated surprising results, proving that the language model can improve itself.\\
\\
...\\
\\
Task: \{instruction for target task\}'''
    \end{minipage}
}
    \caption{Prompt used for generating instances given an instruction. It consists of a constant set of in-context examples. The model has to generate the instances for the given instruction following the format and idea of the given demonstrations.}
    \label{tab:instance_generation_template}
\end{table*}

\subsection{Human Evaluation Details}

We followed the method used by \cite{wang2023selfinstruct} for the human evaluation of 252 user-oriented instructions.

Table \ref{tab:humaneval} presents examples of human annotation. Each entry includes the instruction, input, and target response, as well as the responses generated by both SELF-INSTRUCT and \ourmethod, accompanied by the ratings assigned by evaluators. Notably, our method enhances the understandability and accuracy of model outputs compared to SELF-INSTRUCT. However, it's worth noting occasional instances of model hallucination.




\begin{landscape}
\begin{table}[]
\centering
\begin{tabular}{lllll}
\cline{1-4}
Instruction + Instance Input                                                                                                                                                                                                                                                                                                                                                                                                                   & Target Response                                                                                                                                                                                                                                                                                                                                                                                                                                                                            & SELF INSTRUCT                                                                                                                                                                                                                                                                                                                                         & REFINE AF                                                                                                                                                                                                                                                                                                                                                                                &  \\ \cline{1-4}
\begin{tabular}[c]{@{}l@{}}Instruction: Classify the tweet into one of the \\ following three categories: event, demo, or news.\\ Input: Using this tutorial, you can analyze and \\ classify a user’s review as either positive or \\ negative using sentiment analysis features in an \\ existing Next.js app. We show you how to \\ leverage language AI into your applications.\end{tabular}                                               & Demo                                                                                                                                                                                                                                                                                                                                                                                                                                                                                       & \begin{tabular}[c]{@{}l@{}}Positive\\ \\ Rating: C\end{tabular}                                                                                                                                                                                                                                                                                       & \begin{tabular}[c]{@{}l@{}}Demo\\ \\ Rating: A\end{tabular}                                                                                                                                                                                                                                                                                                                              &  \\ \cline{1-4}
\begin{tabular}[c]{@{}l@{}}Instruction: Please answer the following \\ question-based on the information provided \\ in the article. \\ Input:  Development studies is an\\ interdisciplinary branch of social science. \\ Development studies are offered as a \\ specialized master's degree in several\\ reputed universities around the world.\\ ....\\ What are some popular careers for students \\ of development studies?\end{tabular} & \begin{tabular}[c]{@{}l@{}}Some popular careers for students of \\ development studies include working for \\ an international organisation such as \\ the United Nations, World Bank, or NGOs, \\ as well as working in media, journalism, \\ private sector development consultancy, \\ or corporate social responsibility.\end{tabular}                                                                                                                                                 & \begin{tabular}[c]{@{}l@{}}Development studies are an \\ interdisciplinary branch of social science. \\ Development studies are offered as a \\ specialized master's degree in several \\ reputed universities around the world. \\ ....\\ \\ Rating: D\end{tabular}                                                                                  & \begin{tabular}[c]{@{}l@{}}Students of development studies\\ often choose careers in international\\ organisations such as the United \\ Nations, World Bank, non-governmental \\ organisations (NGOs), media and \\ journalism houses, private sector \\ development consultancy firms\\ corporate social responsibility \\ (CSR) bodies and research centres.\\ Rating: A\end{tabular} &  \\ \cline{1-4}
\begin{tabular}[c]{@{}l@{}}Instruction: Prepare an email signature template\\ for the given person or company.\\ Input:  a graduate student of computer science\end{tabular}                                                                                                                                                                                                                                                                   & \begin{tabular}[c]{@{}l@{}}{[}Your name{]}\\ {[}Your degree{]} in Computer Science\\ {[}Your university{]}\\ Website: {[}Your website{]}\\ ORCID ID: {[}Your ORCID ID{]}\end{tabular}                                                                                                                                                                                                                                                                                                      & \begin{tabular}[c]{@{}l@{}}Hi,\\ I am a graduate student of computer \\ science at {[}University{]}. I am currently \\ working on a project that involves {[}topic{]}. \\ I am looking for a mentor who can help \\ me with this project. If you are interested, \\ please {[}link to a page{]}.\\ Thanks,\\ {[}Your Name{]}\\ Rating: C\end{tabular} & \begin{tabular}[c]{@{}l@{}}{[}Your Email{]}\\ Phone: {[}Your Phone Number{]}\\ LinkedIn: {[}Your LinkedIn URL{]}\\ Twitter: {[}Your Twitter URL{]}\\ Facebook: {[}Your Facebook URL{]}\\ Website: {[}Your Website URL{]}\\ \\ Rating: B\end{tabular}                                                                                                                                     &  \\ \cline{1-4}
\begin{tabular}[c]{@{}l@{}}Instruction: We have entered the home supplies\\ budget in the following table. Calculate the last\\ column with the given function and rewrite\\ the completed table.\\ Input: Item, Qty, Price, Subtotal |\\  mango, 5, 500, ? |\\  apple, 4, 150, ? | \\ oil, 2, 1500, ?\end{tabular}                                                                                                                            & \begin{tabular}[c]{@{}l@{}}Item, Qty, Price, Subtotal | \\ mango, 5, 500, 2500 | \\ apple, 4, 150, 600 | \\ oil, 2, 1500, 3000\end{tabular}                                                                                                                                                                                                                                                                                                                                                & \begin{tabular}[c]{@{}l@{}}mango, 500, 2500 | \\ apple, 150, 3000 | \\ oil, 3000, 45000\\ \\ Rating: C\end{tabular}                                                                                                                                                                                                                                   & \begin{tabular}[c]{@{}l@{}}mango, 5, 500, 2500\\ apple, 4, 150, 600\\ oil, 2, 1500, 3000\\ \\ Rating: A\end{tabular}                                                                                                                                                                                                                                                                     &  \\ \cline{1-4}
\begin{tabular}[c]{@{}l@{}}Instruction: Create a table listing all games that meet \\ the specified criteria in the National Football League. \\ Use the season, local time, game, and score as \\ columns of  the table\\ Input:  Ravens home games in 2011\end{tabular}                                                                                                                                                                      & \begin{tabular}[c]{@{}l@{}}season | local time | game | score\\ 2011-12 | Sun, Jan 15 | Texans (11-7) at...\\ 2011-12 | Sat, Dec 24 | Browns (4-11) at ..\\ 2011-12 | Sun, Dec 11 | Colts (0-13) at ..\\ 2011-12 | Thu, Nov 24 | 49ers (9-2) at...\\ 2011-12 | Sun, Nov 20 | Bengals (6-4)..\\ 2011-12 | Sun, Oct 30 | Cardinals (1-6) ...\\ 2011-12 | Sun, Oct 16 | Texans (3-3) at ...\\ 2011-12 | Sun, Oct 2 | Jets (2-2) at...\\ 2011-12 | Sun, Sep 11 | Steelers (0-1)..\end{tabular} & \begin{tabular}[c]{@{}l@{}}Season: 2011\\ Local time: 12:00 PM\\ Game: Ravens vs. Steelers\\ Score: Ravens 23, Steelers 20\\ \\ Rating: C\end{tabular}                                                                                                                                                                                                & \begin{tabular}[c]{@{}l@{}}Season, Local Time, Game, Score\\ 2011, 17:30, Ravens vs. Browns,\\ Ravens 24, Browns 10\\ 2011, 17:30, Ravens vs. Steelers, \\ Ravens 13, Steelers 10\\ 2011, 17:30, Ravens vs. Bengals,\\ Ravens 23, Bengals 14\\ 2011, 17:30, Ravens vs....\\ \\ Rating: B\end{tabular}                                                                                    &  \\ \cline{1-4}
\end{tabular}
\caption{Examples of Human Evaluation Ratings. Responses for SELF INSTRUCT and \ourmethod\ are given along with their ratings for LLaMa 2 7B.}
\label{tab:humaneval}
\end{table}

\end{landscape}

\bibliographystyle{IEEEtran}
\bibliography{main}

\end{document}